\documentclass{vgtc}                          

\graphicspath{{figures/}{pictures/}{images/}{./}} 

\usepackage{times} 
\usepackage{amsmath}
\usepackage{multirow}

\usepackage{tabu}                      
\usepackage{booktabs}                  
\usepackage{lipsum}                    
\usepackage{mwe}                       
\usepackage{colortbl}
\usepackage{float} 
\usepackage{caption}

\usepackage{mathptmx}                  

\onlineid{1134}

\vgtccategory{Research}

\vgtcinsertpkg

\title{EgoCHARM: Resource-Efficient Hierarchical Activity Recognition using an Egocentric IMU Sensor}

\author{Akhil Padmanabha\thanks{Work done while at Meta. email: akhil.padmanabha@gmail.com}\\ %
        \scriptsize Carnegie Mellon University \vspace{1em}%
\and Saravanan Govindarajan\\ %
     \scriptsize Meta Reality Labs %
\and Hwanmun Kim\\ %
     \scriptsize Meta Reality Labs %
\and Sergio Ortiz\\ %
     \scriptsize Meta Reality Labs %
\and Rahul Rajan\\ %
     \scriptsize Meta Reality Labs %
\and Doruk Senkal\\ %
     \scriptsize Meta Reality Labs %
\and Sneha Kadetotad\\ %
     \scriptsize Meta Reality Labs%
}

\teaser{
  \centering
  \includegraphics[width=\linewidth]{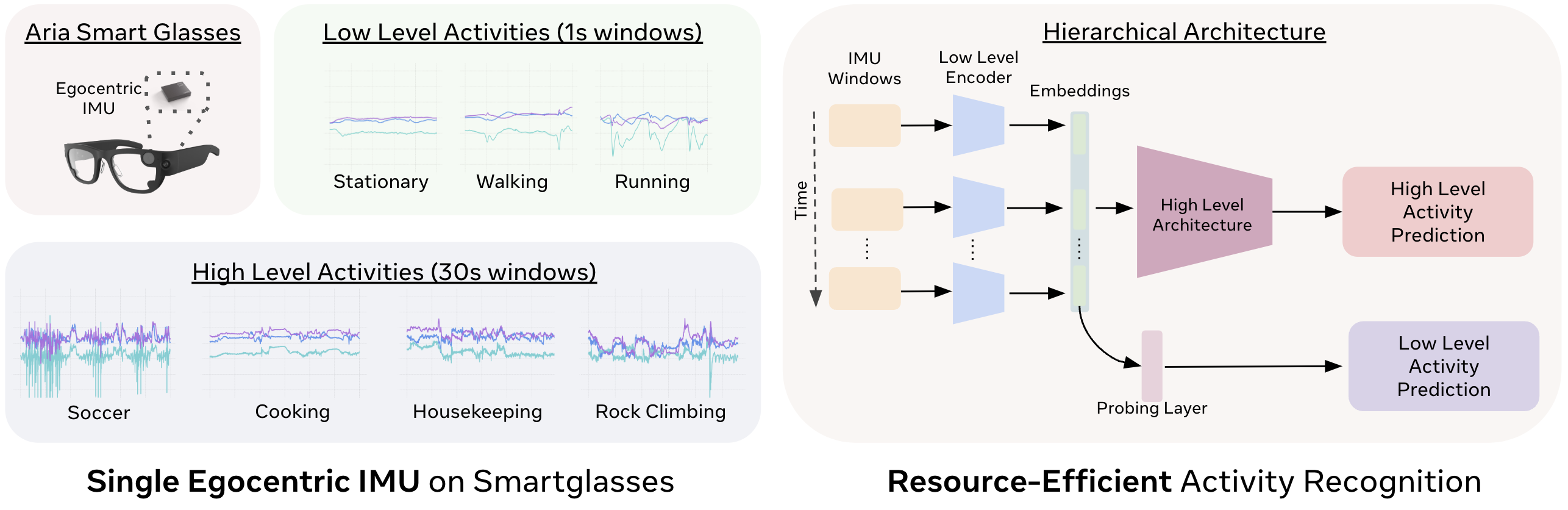}
  \caption{We propose a resource-efficient hierarchical architecture, EgoCHARM, to classify both high and low level activities using a single egocentric, head-mounted IMU. On the left, we show the Aria smartglasses~\cite{engel2023project,projectaria} featuring the IMU along with sample acceleration data from selected low and high level activities within our dataset. Our hierarchical architecture, shown on the right, features a low level encoder that inputs 1s low level windows of IMU data and extracts motion embeddings, which can be used to predict low level activities or aggregated over time (30s) and inputted into a high level architecture to predict high level activities.}
  \label{fig:teaser}
}

\abstract{
   Human activity recognition (HAR) on smartglasses has various use cases, including health/fitness tracking and input for context-aware AI assistants. However, current approaches for egocentric activity recognition suffer from low performance or are resource-intensive. In this work, we introduce a resource (memory, compute, power, sample) efficient machine learning algorithm, EgoCHARM, for recognizing both high level and low level activities using a single egocentric (head-mounted) Inertial Measurement Unit (IMU). Our hierarchical algorithm employs a semi-supervised learning strategy, requiring primarily high level activity labels for training, to learn generalizable low level motion embeddings that can be effectively utilized for low level activity recognition. We evaluate our method on 9 high level and 3 low level activities achieving 0.826 and 0.855 F1 scores on high level and low level activity recognition respectively, with just 63k high level and 22k low level model parameters, allowing the low level encoder to be deployed directly on current IMU chips with compute. Lastly, we present results and insights from a sensitivity analysis and highlight the opportunities and limitations of activity recognition using egocentric IMUs.
} 

\keywords{wearables, human activity recognition, motion sensors, inertial measurement units}

\begin{document}


\firstsection{Introduction}

\maketitle

The proliferation of wearable devices and sensor-enabled technologies in portable form factors has created numerous opportunities for tracking, analyzing, and generating insights into human actions and behaviors. Human activity recognition (HAR), the process of automatically identifying and classifying an individual's physical activities from sensor data, has become a critical component in various consumer applications, with many existing and potential future use cases including generating daily activity timelines, tracking health/fitness, and providing personalized insights and recommendations. Head-mounted devices such as smartglasses present a unique opportunity for activity recognition by providing a first-person view of the world from the user's perspective. While numerous studies have demonstrated activity recognition using egocentric cameras in smartglasses~\cite{grauman2024ego, liu2022egocentric, ma2016going, grauman2022ego4d, song2023ego4d, nguyen2016recognition}, there are significant hurdles to overcome before always-on vision algorithms can be deployed in real-world settings, including high power consumption, substantial computational demands, and privacy concerns.

In comparison, Inertial Measurement Units (IMUs), which capture human motion through acceleration and angular velocity signals, are well-suited for continuous activity recognition due to their low power consumption, minimal computational requirements, and inherent privacy advantages. Additionally, IMUs are increasingly being equipped with on-chip compute capabilities, enabling further resource savings on the primary device processor~\cite{BoschSensortec2023BHI360, STMICRO}. In addition to activity recognition use cases, always-on IMU sensing provides strong advantages for compact form factors like smartglasses, including the ability to trigger higher-power sensors for other functionalities on the device, enabling more efficient use of system resources. Furthermore, motion embeddings from IMU activity recognition algorithms can be used as input for large language models, unlocking further capabilities like context-aware AI assistants without the need for explicit prompting using language~\cite{moon2022imu2clip, li2024sensorllm, moon2024anymal}. The emergence of consumer smartglasses, such as the Meta Ray-Bans, motivates the use of head-mounted IMUs for always-on activity recognition on smartglasses. In comparison to body and wrist-mounted IMUs, which have been researched extensively for activity recognition, signals from head-mounted IMUs are distinct and remain relatively unexplored in the literature. 

\begin{figure}[htp]
    \centering
    \includegraphics[width=\columnwidth]{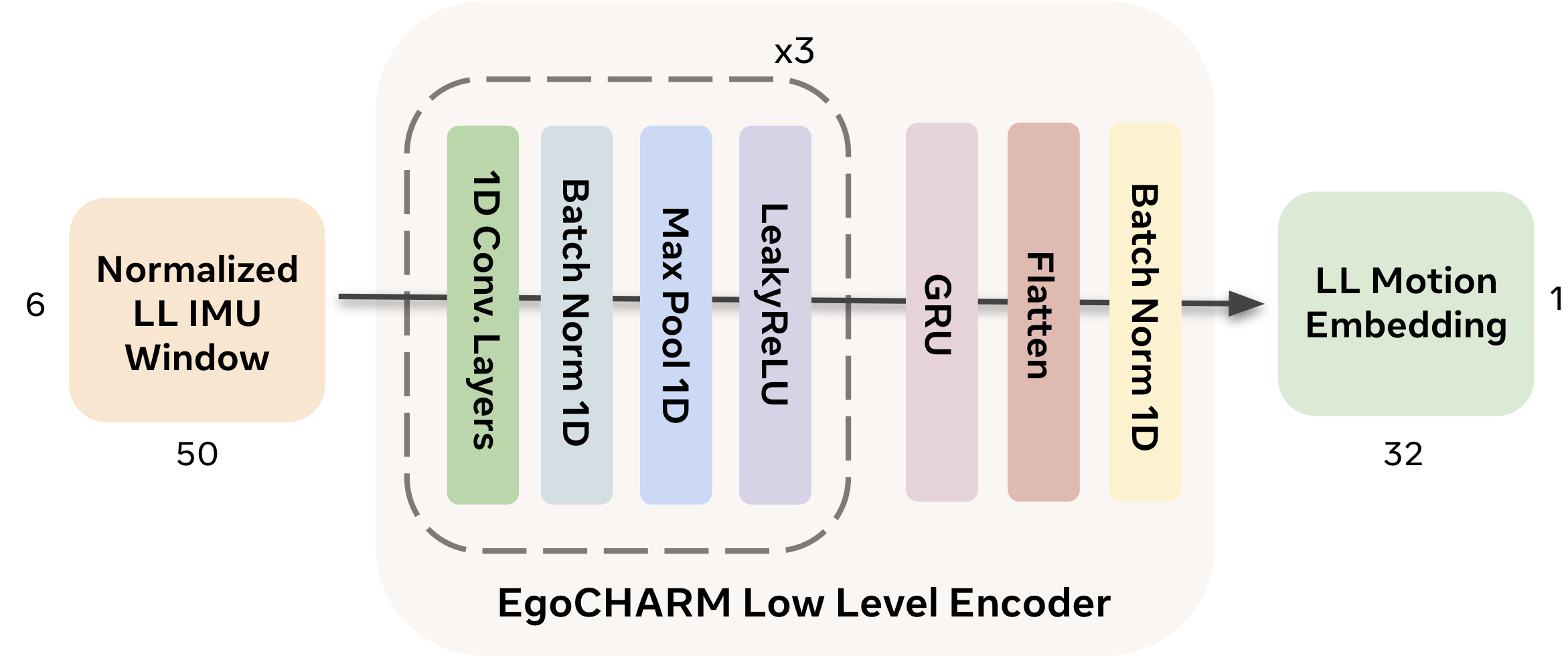}
    \caption{EgoCHARM Low Level Encoder Architecture. Our encoder consists of 1D-CNN (Convolutional Neural Network) layers with variable dilation to capture the periodic patterns present in IMU signals and a GRU (Gated Recurrent Unit) for capturing temporal sequences.}
    \label{fig:ll_encoder}
\end{figure}

In this work, we introduce a resource (memory, compute, power, sample) efficient hierarchical machine learning algorithm, EgoCHARM, for recognizing both high level and low level activities using a single egocentric IMU on smartglasses. We define high level activities as those lasting more than 30 seconds, including tasks like fitness/exercise (e.g., basketball, soccer), chores (e.g., cooking, cleaning), and prolonged stationary activities (e.g., working on a laptop, using a phone). Conversely, low level activities are brief, simple actions occurring within a 1 second window, such as walking and remaining stationary. Low level activities can be aggregated to form the more complex high level activities, motivating the use of a hierarchical framework. As illustrated in Fig.~\ref{fig:teaser}, our hierarchical machine learning algorithm features a novel low level encoder, a CNN-GRU (Convolutional Neural Network-Gated Recurrent Unit) architecture shown in Fig.~\ref{fig:ll_encoder}, with a low parameter count, allowing it to be deployed directly on existing IMU chips, saving compute and memory on the main device processor. This encoder, trained in a semi-supervised method using only high level activity labels, extracts latent motion embeddings over short time horizons. These embeddings are then inputted sequentially into our high level architecture, a GRU (Gated Recurrent Unit) architecture, which utilizes a longer time horizon to classify complex high level activities. We show how this low level encoder can be frozen and probed to distinguish between three low level activities (stationary, walking, running), highlighting the utility and generalizability of the low level embeddings. With the EgoCHARM architecture, we demonstrate strong performance in both high level and low level activity recognition achieving 0.826 test F1 score on 9 high level classes and 0.855 test F1 score on 3 low level classes using a low number of parameters (22k low level and 63k high level). By classifying both high and low level activities, we can gain a comprehensive understanding of an individual's daily routine, enabling applications in health monitoring, activity tracking, and personalized recommendations that require a nuanced view of both short-term actions and long-term behaviors. Our work additionally provides insights from our findings including a sensitivity analysis highlighting the effect of various parameters including number of samples and IMU sampling frequency on performance. Lastly, we present challenges of using egocentric IMUs for activity recognition and future opportunities in this field. 

In summary, the contributions of this work are as follows:
\begin{itemize}
    \item We present a resource-efficient hierarchical machine learning algorithm, EgoCHARM, for classification of high level activities using a single egocentric IMU on smartglasses, achieving 0.826 test F1 score and 82.86\% test accuracy for 9 class high level activity recognition. 
    \item We show the generalizability of EgoCHARM's semi-supervised low level encoder through probing, achieving 0.855 test F1 score and 90.64\% test accuracy for classification between 3 low level activities.
    \item We present results and insights from a sensitivity analysis and highlight the opportunities and limitations of activity recognition and generalizable encoders using egocentric, head-mounted IMUs. 
\end{itemize}

\begin{table*}[htbp] 
    \centering
    \caption{High and Low Level Dataset Details}
    \vspace{-1em}
    \label{tab:dataset_info}
    \begin{tabular}{|c|c|c|c|c|c|}
        \toprule
        \textbf{Dataset} & \textbf{Classes} & \textbf{Window Size (s)} & \textbf{Stride (s)} & \textbf{Train Samples} & \textbf{Test Samples} \\
        \midrule
        High Level & 
        \begin{tabular}{@{}c@{}}
            9: (soccer, basketball, dance, rock climbing, \\ body stretch, housekeeping, cooking, 
            bike repair, music)
        \end{tabular} 
        & 30 & 10 & 48,078 & 16,063 \\
        \midrule
        Low Level & 3: (stationary, walking, running) & 1 & 1 & 14,343 & 4,930 \\
        \bottomrule
    \end{tabular}
\end{table*}

\section{Related Works}
The field of human activity recognition has seen numerous advancements through the use of various devices and sensing modalities including cameras and IMUs~\cite{chen2021deep,lara2012survey, ke2013review, vrigkas2015review}. In this section, we focus on related work pertinent to two topics that our work intersects: egocentric (head-mounted) IMU activity recognition and resource-efficient IMU activity recognition. 

\subsection{Egocentric IMU Activity Recognition}
 Egocentric IMUs have been used on head-mounted devices such as smartglasses and AR/VR headsets for applications including indoor localization / SLAM~\cite{lv2024aria, bamdad2024incrowd}, head pose and gesture estimation~\cite{severin2020head, kumar2023comparison, li2024head}, and as input interfaces for devices and robots~\cite{padmanabha2023hat, padmanabha2024independence, padmanabha2025towards}. We focus on research related to egocentric IMUs, as the signals from head-mounted sensors differ from those obtained from body-mounted sensors affecting activity classification performance. Most existing works in this area lack generalizability across participants or rely on small datasets collected under controlled conditions with limited activities and/or participants~\cite{gjoreski2021head, cristiano2019daily, ionut2021using}. Other studies have explored pretraining of generalizable egocentric IMU encoders for various downstream activity recognition tasks~\cite{moon2022imu2clip, das2024primus, chen2025comodo}. Notable methods include IMU2CLIP~\cite{moon2022imu2clip} and PRIMUS~\cite{das2024primus}, but these suffer from suboptimal performance on activity recognition while utilizing large architecture sizes. For example, PRIMUS employs a stacked RNN architecture with approximately 1.4M parameters, achieving around 70\% accuracy on 8 classes from the Ego-Exo4D dataset~\cite{das2024primus}. In contrast, EgoCHARM is more compact, demonstrates strong performance on high level activities, and can generalize to predict low level activities.

\subsection{Resource-Efficient IMU Activity Recognition}
\label{sec:related_work_resource_efficient}
Minimizing model parameters, FLOPs, and IMU sampling frequencies is crucial for enabling deployment of IMU activity recognition on resource-constrained devices like smartglasses. In practice, many consumer devices and IMUs with on-chip compute like the Bosch BMI263~\cite{BoschSensortec2023BMI263} use decision trees for simple activity recognition as they are require low compute and memory~\cite{jaimes2016pat, fan2013human}. However, performance with decision trees is limited and thus, many recent works have explored neural network architectures to address this gap while still considering compute and memory constraints on the edge~\cite{nakabayashi2024multimodal, ravi2016deep, rosen2022charm}. However, no prior works explore resource-efficient activity recognition algorithms with neural network architectures using egocentric IMUs. 

Hierarchical architectures~\cite{rosen2022charm, lemieux2020hierarchical, yu2019hierarchical} present considerable promise for activity recognition, as they inherently capture the hierarchy between low level and high level activities, with low level activities collectively forming high level ones. Furthermore, these architectures are inherently more resource-efficient as they can be easily divided, allowing the more frequently executed low level architecture to be run directly on the IMU chip, saving compute and memory on the main device processor. Our work specifically draws inspiration from CHARM (Complex Human Activity Recognition Model)~\cite{rosen2022charm}, a hierarchical deep neural network architecture that leverages IMU data to classify high level activities by learning low level motion patterns. A key benefit of CHARM is its ability to train with only high level labels, reducing the need for extensive annotation efforts in dataset creation. However, CHARM's evaluation was limited to multiple body-mounted IMUs and a small dataset with few activities and participants. Additionally, it used hand-picked IMU features (mean, minimum, maximum, etc.) and multi-layer perceptrons for both low and high level architectures, limiting performance. Our work builds upon and expands hierarchical CHARM models by exploring more efficient and effective model architectures, evaluating on a large dataset with many participants, and demonstrating generalizability of the semi-supervised low level IMU encoders for low level activity recognition. 

\section{Methods}
We frame both high and low level activity recognition as supervised learning classification problems, where the goal is to categorize each high level sample into one of $n$ distinct classes.

\subsection{Datasets}
\label{sec:datasets}

We utilize IMU (3-axis accelerometer and 3-axis gyroscope) data from two publicly available datasets, Ego-Exo4D~\cite{grauman2024ego} and Nymeria~\cite{ma2024nymeria}. Ego-Exo4D includes data gathered from 740 participants, whereas Nymeria comprises data from 264 participants. Both datasets were collected using the Aria V1 smartglasses from Meta Reality Labs~\cite{engel2023project, projectaria}. The Aria smartglasses consist of multiple sensors including cameras and two IMUs, on the left and right sides of the device. We use data from only the left IMU and downsample to 50 Hz. 

For high level activity recognition, we use 9 total classes, 7 from Ego-Exo4D and 2 from Nymeria. We use the ``scenario'' annotations from both datasets to select our high level activities. Specifically from the Ego-Exo4D dataset, we use the basketball, soccer, cooking, rock climbing, dance, and bike repair scenarios as our high level classes. From the Nymeria dataset, we use the body stretch and housekeeping scenarios as our high level classes. For both datasets, we exclude the rest of the scenarios because they encompass numerous sub-high level activities with different motion profiles. For example, in the Ego-Exo4D dataset, the health scenario includes both administering a COVID-19 test and performing CPR. We show sample IMU data from multiple high level activities in Fig.~\ref{fig:teaser}, Fig.~\ref{fig:cooking}, and Fig.~\ref{fig:bike_repair}.

For low level activity recognition, we use 3 total classes: stationary, walking, and running and we select this data from the Nymeria dataset. For reannotation, we use the ``activity summarization" annotations, provided for every 30 seconds of data, and manually select windows of data that match our low level classes using the textual summaries. The majority of the stationary samples are from the ``by my desk'' scenario,  the majority of the walking samples are from the ``hiking'' scenario, and the majority of the running samples are from the ``fresh air'' scenario. For the running class, we additionally refine our annotations visually from the RGB camera footage from the Aria glasses. We show sample IMU data from the three low level activities in Fig.~\ref{fig:teaser}.

We split both our high and low level datasets into a train and test set using stratification, ensuring that no participant's data exists in both the train and test sets, while preserving the percentage of samples for each class in both sets. We use high level window sizes of 30 seconds and low level window sizes of 1 second. These window sizes are consistent with typical annotation schemes used in egocentric datasets, where low level activities, commonly called ``atomic actions'', are annotated within 1 second windows and high level activities, such as ``activity summarization'' from the Nymeria dataset, are annotated over 30 second windows~\cite{grauman2024ego, grauman2022ego4d, ma2024nymeria}. The amount of samples in our datasets is shown in Table~\ref{tab:dataset_info}. Fig.~\ref{fig:hl_samples} and Fig.~\ref{fig:ll_samples} additionally show the number of samples per class in each dataset. We ensure that samples used in our low level dataset are excluded from our high level dataset.

\subsection{Hierarchical Modeling}
\label{sec:hierarchical_architecture}
As discussed in Section~\ref{sec:related_work_resource_efficient}, we draw inspiration from the hierarchical model, CHARM, proposed by Rosen et al.~\cite{rosen2022charm}. Our algorithm, EgoCHARM, is shown visually in Fig.~\ref{fig:teaser} and Fig.~\ref{fig:ll_encoder}, and we additionally describe it mathematically as follows. For a high level data window spanning \( t_{HL} = 30 \) seconds, we initially downsample each of the 6 IMU channels to 50 Hz using linear interpolation. We specifically use 50 Hz as according to the Nyquist theorem, it enables us to accurately capture signals up to 25 Hz, which encompasses most human movements. We then partition the high level window into non-overlapping low level windows, each spanning \( t_{LL} = 1 \) second. For specific model architectures explored, discussed in further detail Section~\ref{sec:low_level_encoder_architectures}, we extract hand picked features per channel while for the majority of the model architectures, we directly input the 6 channels of raw data. While using hand-picked features, we flatten and normalize these features to zero mean and unit variance. For the hand-picked features, the low level input to our low level encoder is \(\mathbf{f_{LL}} \in {R}^{30}\), as there are 6 IMU channels with 5 features each.  For the raw data architectures, we apply normalization across all windows of raw data, specifically along each of the 6 IMU channels. The input is defined as \(\mathbf{f_{LL}} \in {R}^{6 \times 50}\), as we use 1 second low level windows with a sampling rate of 50 Hz. The low level input, \(\mathbf{f_{LL}}\), is fed into our low level encoder, \text{LLE}, which outputs a low level motion embedding, \(\mathbf{e_{LL}}\), for each 1s low level window of data, \(\mathbf{e_{LL}} = \text{LLE}(\mathbf{f_{LL}})\) where \(\mathbf{e_{LL}} \in {R}^{o}\), where $o$ is the output embedding dimension. 

For high level activity recognition, we concatenate the low level output embeddings across the entire 30s high level window to form a concatenated embedding vector, \(\mathbf{e_{HL}} \in {R}^{n*o}\), where $n$ is the number of low level windows in a high level window; in this case, $n=30$. This vector is then inputted into a high level architecture, \text{HLA}. A softmax activation function is applied to obtain classification probabilities. This can be expressed as \(\mathbf{p_h} = \text{HLA}(\mathbf{e_{HL}})\) where \(\mathbf{p_h}\) is the output vector containing the high level classification probabilities. We train both our low level encoder and high level architecture concurrently using only high level activity labels and a weighted cross-entropy loss to handle class imbalance. We specify how we probe our trained low level encoder to enable low level activity recognition, in Section~\ref{sec:probing}. 

\subsection{Low Level Encoder Architectures}
\label{sec:low_level_encoder_architectures}

In designing our low level encoder, we focus on developing small model architectures with fewer than  25,000 parameters, driven by the on-chip compute capabilities of IMUs available today. Commercially available IMU chips, such as the Bosch BHI360~\cite{BoschSensortec2023BHI360} and STMicroelectronics LSM6DSO16IS~\cite{STMICRO}, typically feature on-chip compute resources ranging from 32 kB to 256 kB of program memory, with a portion occupied by manufacturer tooling, code, and other components. By leveraging quantization and other optimization techniques, we can deploy our low level encoders on these resource-constrained devices, as long as they remain within the 25,000 parameter threshold. We explored the following architectures:
\begin{itemize}
    \item MLP with Hand-Picked Features: Inspired by Rosen et al.~\cite{rosen2022charm}, for each of the low level raw data windows, we extract 5 hand-picked features (mean, maximum, minimum, variance, and peak to peak difference). These features are inputted into an MLP architecture. 
    \item CNN: We utilize 1-D CNN architectures where convolutions are applied over each of the 6 IMU channels. At each convolutional layer, multiple convolutions with the same kernel size and different dilations are applied in parallel and stacked together to capture the periodicity of the IMU signals.
    \item IMU2CLIP: Both IMU2CLIP~\cite{moon2022imu2clip} and PRIMUS~\cite{das2024primus} use a CNN+GRU architecture with group norm and fixed kernel size and dilation for the CNN layers. Due to the IMU2CLIP encoder having over 1 million parameters, we explore the same architecture but with smaller model dimensions. In line with the original work, we do not normalize our IMU data before inputting it into the encoder.
    \item CNN-LSTM (Long Short-Term Memory): We use 1-D CNN layers with variable dilation as discussed for the CNN method. We apply an LSTM on the output. The encoder outputs the last output of the final LSTM layer. 
    \item CNN-GRU: We use 1-D CNN layers with variable dilation as described above with a GRU applied on the output. The encoder outputs the last output of the final GRU layer. This architecture is shown visually in Fig.~\ref{fig:ll_encoder}.
\end{itemize}

\begin{figure}[htbp]
    \centering
    \includegraphics[width=\columnwidth]{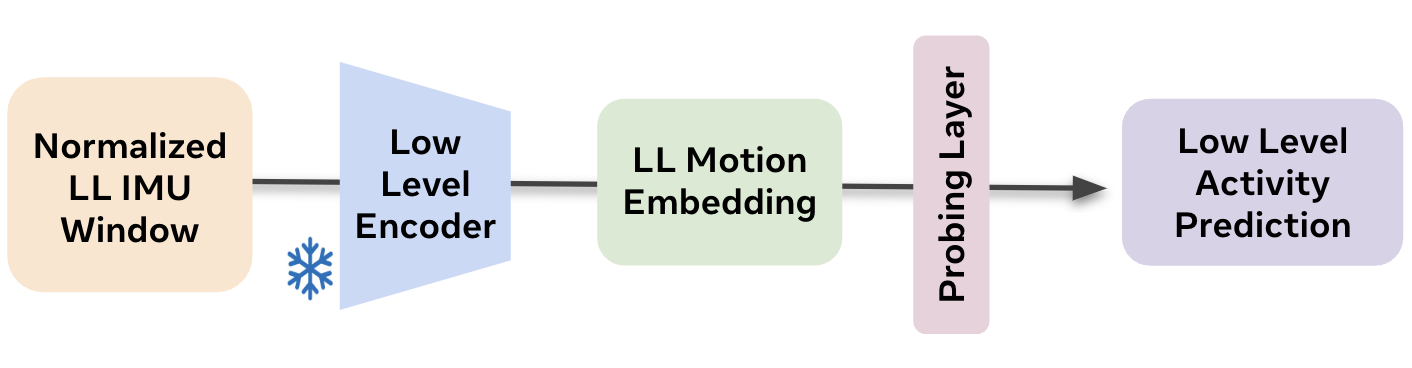}
    \caption{Low Level Encoder Probing. To enable low level activity recognition, we freeze our low level encoder's parameters and train a probing layer to map our low level motion embeddings to 3 discrete classes.}
    \label{fig:probing}
\end{figure}

\subsection{High Level Architectures}
\label{sec:high_level_architectures}

As high level architectures will run on the main device processor, we can be more flexible on memory/compute requirements, but we still aim to find small models without sacrificing performance. We explored the following high level architectures:
\begin{itemize}
    \item MLP: We concatenate the LLE output embeddings over the entire high level window and input these into a MLP architecture for high level classification. 
    \item GRU: We input the LLE output embeddings over the entire high level window sequentially through a GRU architecture. The last output of the GRU is used for the high level classification.
    \item LSTM: We input the LLE output embeddings over the entire high level window sequentially through a LSTM architecture. The last output of the LSTM is used for the high level classification.
\end{itemize}

\subsection{Clustering on Low Level Motion Embeddings}
\label{sec:pca}
To evaluate the generalizability of our low level encoder, which was trained using only high level labels, we employ principal component analysis (PCA) to cluster output vectors in the latent space. We utilized 80 randomly selected samples per class (stationary, walking, and running) from our low level activity dataset. We used our trained EgoCHARM low level encoder, shown in Fig.~\ref{fig:ll_encoder}, to extract embeddings, each with a dimension of 1 by 32. Subsequently, we applied PCA with 2 dimensions to these embeddings and visualize the results to provide insight on the generalizability of our low level encoder to unseen low level classes. 

\subsection{Low Level Encoder Probing}
\label{sec:probing}
The EgoCHARM hierarchical architecture allows our low level encoder to automatically learn low level motion embeddings using only high level labels in a semi-supervised fashion. To enable low level activity recognition, we must map these low level motion embeddings to discrete low level classes. We frame this as a supervised learning problem. 

In line with past work on evaluating generalizability of IMU encodings for activity recognition~\cite{moon2022imu2clip, das2024primus}, we freeze our low level encoder's parameters and add a leaky ReLU activation function, a single probing layer (PL), and softmax. This is shown visually in Fig.~\ref{fig:probing}. The probing layer's weights are tunable and are randomly initialized. This step can be expressed as $\mathbf{p_{LL}} = \text{PL} ( \text{LLE}_f(\mathbf{f_{LL}}))$ where $\text{PL}$ is the probing layer, $\text{LLE}_f$ is the frozen low level encoder, and $\mathbf{p_{LL}}$ is the output vector containing the low level classification probabilities.

We utilize our low level dataset introduced in Section~\ref{sec:datasets} to train our probing layer on the 3 low level activities. We then evaluate its performance using our held out test set. Due to class imbalance, we apply a weighted cross-entropy loss during training. Additionally, due to the low number of running class samples in our low level dataset, we evaluate performance using 4 fold cross validation using stratification to ensure no participant data is both in the train and test set and to ensure equal percentage of classes/samples across folds.

\subsection{Experiments and Hyperparameter Searches}
\label{sec:methods_sensitivity_analysis}
To determine our best performing model, we run hyperparameter sweeps for high level activity recognition on every combination of low and high level architectures presented in Section~\ref{sec:low_level_encoder_architectures} and Section~\ref{sec:high_level_architectures}. For each architecture combination, we conduct 200 runs and use Sobol sampling~\cite{bergstra2012random} in addition to a Bayesian optimization method, SAASBO~\cite{eriksson2021high}, to select parameters for each run, optimizing over the test F1 score. After selecting the best performing model for every architecture combination, we evaluate the frozen low level encoder's effectiveness at low level activity recognition using our low level activity dataset and probing layer detailed in Section~\ref{sec:probing}. Further details on training and hyperparameter searches are provided in Section~\ref{sup:hyperparam_searches}. 

We additionally conduct a sensitivity analysis to evaluate the impact of various factors on performance, including the number of high level samples per class, the number of low level samples per class, sampling frequency, and high level window size. For the number of HL samples per class, we test 500, 1000, 1500, 2000, and 2500 samples. For LL samples per class, we experiment with 250, 500, 1000, 2000, and 3000 samples. In both cases, we randomly select the specified number of samples from the original training datasets, detailed in Table~\ref{tab:dataset_info}, Fig.~\ref{fig:hl_samples}, and Fig.~\ref{fig:ll_samples}. At higher numbers of samples per class, we exhaust the available samples for some classes. For sampling frequency, we evaluate at 15, 25, 75, and 100 Hz. Lastly, for the high level window size, we test 5, 10, 15, 20, and 25 seconds. All parameters of the EgoCHARM processing pipeline and architecture remain constant, except for the parameter being modified. In line with our other hyperparameter sweeps, we perform 200 runs, only allowing variations in training parameters such as learning rate, number of epochs, etc. Further details are provided in Section~\ref{sup:hyperparam_searches}.

\begin{table*}[ht]
\centering
\caption{Performance, Memory, and Compute Metrics}
\vspace{-1em}
\begin{tabular}{|c||c|c||c|c||c|c|c||c|c|c|}
\toprule
\multicolumn{1}{|c||}{\textbf{Algorithm}} & \multicolumn{2}{c||}{\textbf{F1 Score}} & \multicolumn{2}{c||}{\textbf{Accuracy (\%)}} & \multicolumn{3}{c||}{\textbf{Number of Parameters}} & \multicolumn{3}{c|}{\textbf{FLOPs}} \\
\midrule
\textbf{LL} + \textbf{HL} & \textbf{LL} & \textbf{HL} & \textbf{LL} & \textbf{HL} & \textbf{LL} & \textbf{HL} & \textbf{PL} & \textbf{LL} & \textbf{HL} & \textbf{PL} \\
\toprule
MLP + MLP & 0.712 & 0.773 & 84.56 & 78.47 & 5,408 & 314,121 & 99 & 5,280 & 313,681 & 96 \\
CNN + MLP & 0.729 & 0.767 & 83.67 & 78.08 & 17,656 & 35,529 & 51 & 197,040 & 35,473 & 48 \\
IMU2CLIP + MLP & 0.623 & 0.740 & 71.26 & 75.96 & 27,372 & 191,241 & 51 & 523,584 & 190,801 & 48 \\
CNN-LSTM + MLP & 0.781 & 0.807 & 86.77 & 81.21 & 17,772 & 66,249 & 99 & 837,872 & 66,193 & 96 \\
CNN-GRU + MLP & 0.820 & 0.796 & 89.07 & 81.28 & 15,532 & 66,249 & 99 & 729,072 & 66,193 & 96 \\
\hline
MLP + LSTM & 0.762 & 0.778 & 88.02 & 77.28 & 7,760 & 58,953 & 99 & 7,584 & 1,732,497 & 96 \\
CNN + LSTM & 0.769 & 0.793 & 88.66 & 79.61 & 15,568 & 21,577 & 51 & 162,432 & 620,817 & 48 \\
IMU2CLIP + LSTM & 0.781 & 0.731 & 85.25 & 75.15 & 30,252 & 25,673 & 99 & 548,928 & 743,697 & 96 \\
CNN-LSTM + LSTM & 0.691 & 0.802 & 80.53 & 80.45 & 9,780 & 825,609 & 99 & 472,400 & 24,624,465 & 96 \\
CNN-GRU + LSTM & 0.790 & 0.804 & 88.61 & 80.40 & 4,381 & 825,609 & 99 & 209,700 & 24,624,465 & 96 \\
\hline
MLP + GRU & 0.753 & 0.798 & 87.07 & 80.93 & 7,760 & 225,033 & 99 & 7,584 & 6,660,945 & 96 \\
CNN + GRU & 0.761 & 0.792 & 84.43 & 79.42 & 15,568 & 212,745 & 51 & 162,432 & 6,292,305 & 48 \\
IMU2CLIP + GRU & 0.689 & 0.736 & 79.86 & 75.73 & 30,252 & 19,401 & 99 & 548,928 & 559,377 & 96 \\
CNN-LSTM + GRU & 0.755 & 0.818 & 87.08 & 82.69 & 26,220 & 225,033 & 99 & 1,252,272 & 6,660,945 & 96 \\
\arrayrulecolor{lightgray}
\hline
\rowcolor{lightgray}
\textbf{EgoCHARM: CNN-GRU + GRU} & \textbf{0.855} & \textbf{0.826} & \textbf{90.64} & \textbf{82.86} & 21,868 & 63,369 & 99 & 1,041,072 & 1,855,953 & 96 \\
\bottomrule
\end{tabular}
\label{tab:results_table}
\caption*{Note: For high level activity recognition, hyperparameter sweeps (200 runs) were conducted for every combination of low and high level architectures presented in the above table. For low level activity recognition, we conduct 4 fold cross validation and average results across folds.}
\end{table*}

\section{Results}

\begin{figure}[htbp]
    \centering
    \includegraphics[width=0.9\columnwidth]{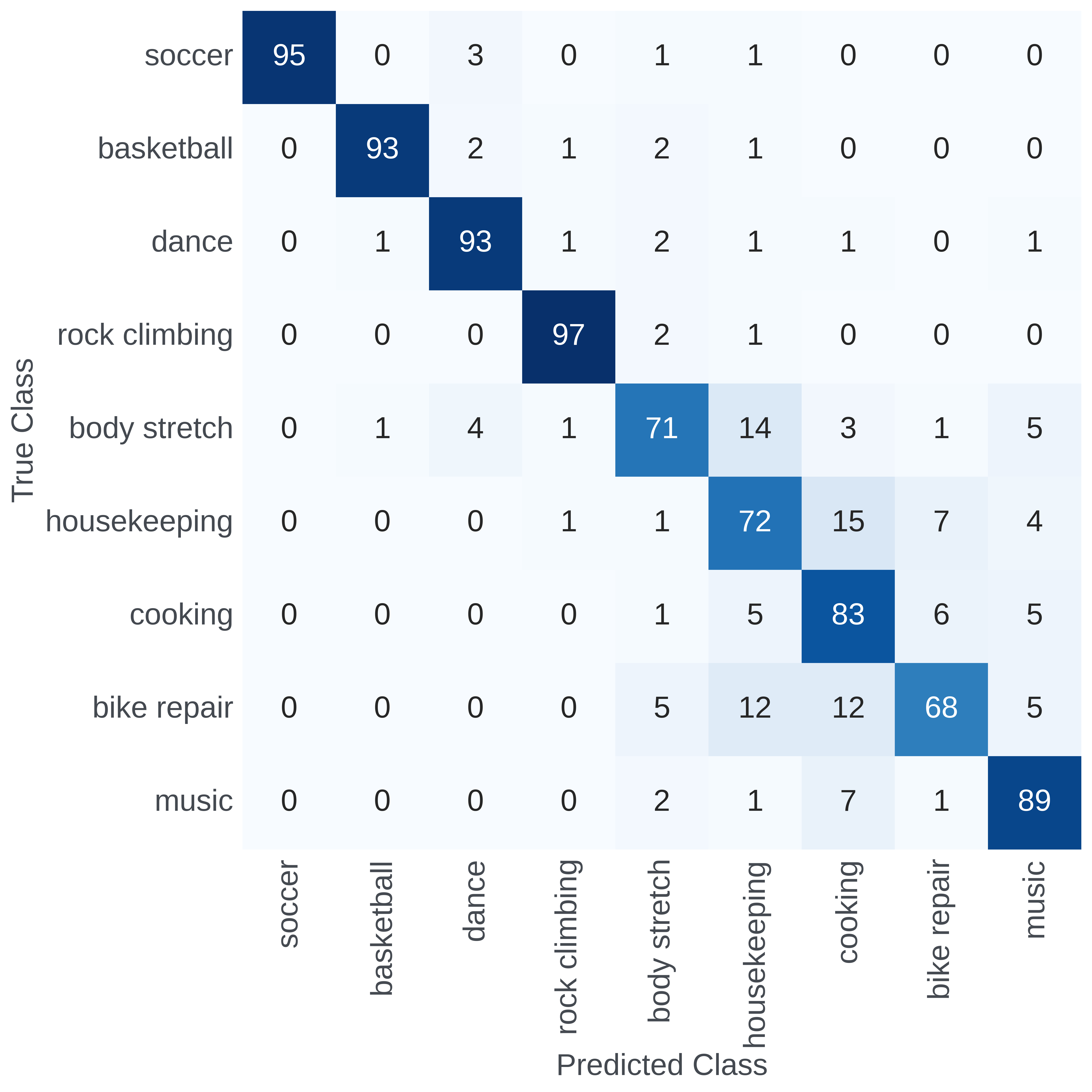}
    \caption{High Level Activity Recognition Confusion Matrix using EgoCHARM. Values in the confusion matrix are in percentage.}
    \label{fig:hl_cm}
\end{figure}

Table~\ref{tab:results_table} shows performance and resource metrics for all models evaluated for both high and level activity recognition. For reported test F1 scores, we use macro averaging while for test accuracy we use micro averaging. Macro averaging calculates the metric for each class separately and then computes the average of these individual metrics while micro averaging calculates the metric by considering all samples together. To benchmark memory, we count the number of parameters in our model architecture, whereas for benchmarking computational performance, we use FLOPs, which are the total number of floating point operations required for a forward pass of the model. We acknowledge that FLOPs is not an ideal metric for benchmarking compute because the required compute is highly dependent on the chip architecture and specific optimizations. Nevertheless, it provides a general indication of the computational demands of a model, serving as a baseline for comparison.

\subsection{EgoCHARM Architecture and Performance}
 As seen in Table~\ref{tab:results_table}, the EgoCHARM architecture obtains the highest classification performance out of all models evaluated with a test F1 score of 0.826 and a test accuracy of 82.86\%. EgoCHARM, consists of a CNN-GRU low level encoder, shown in Fig.~\ref{fig:ll_encoder}, and a GRU high level architecture. The architecture has only 21,868 low level parameters and 63,369 high level parameters, with 1,041,072 low level FLOPs and 1,855,953 high level FLOPs. As seen in the confusion matrix in Fig.~\ref{fig:hl_cm}, the model achieves strong performance for most of the classes, especially higher motion classes including soccer, basketball, dance. Misclassifications are primarily confined to the classes with manipulation of objects like housekeeping, cooking, bike repair, and music. 

\begin{figure}[htbp]
    \centering
    \includegraphics[width=\columnwidth]{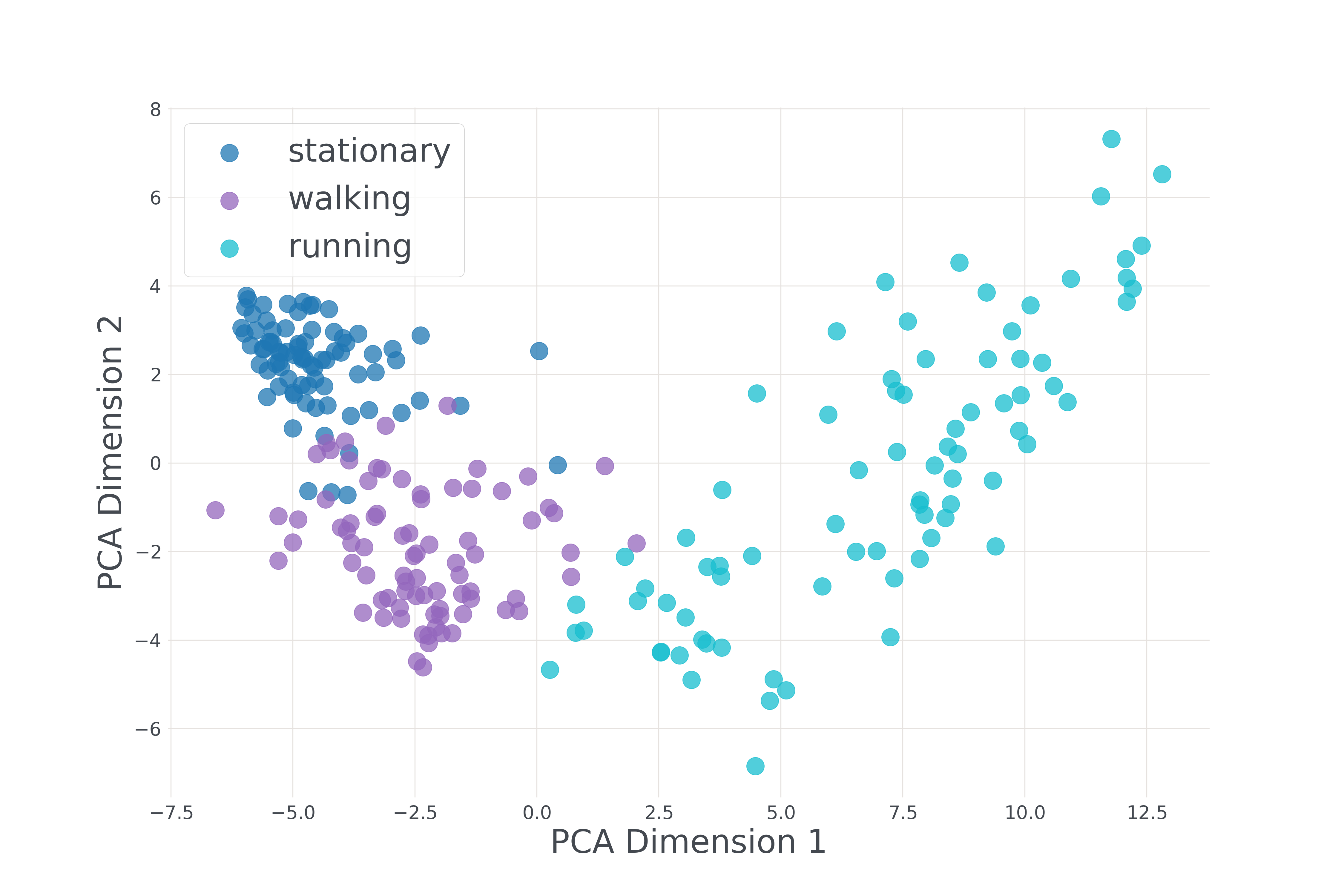}
    \caption{Principal component analysis (PCA) on unseen low level activity motion embeddings using EgoCHARM low level encoder trained using only high level activity labels. Distinct clusters are seen for the three low level classes.}
    \label{fig:pca_LL}
\end{figure}

As detailed in Section~\ref{sec:pca}, we use PCA to cluster on the low level embeddings from our semi-supervised low level encoder in order to evaluate the generalizability to completely unseen low level data. As seen in Fig.~\ref{fig:pca_LL},  despite not being trained on low level labels, our EgoCHARM low level encoder distinctly clusters the 3 activities. Notably, clusters of more closely related activities (e.g., stationary/walking and walking/running) are positioned adjacently with slight overlap, while the classes with more distinct motion profiles (stationary/running) are distanced farther away.

As discussed in Section~\ref{sec:probing}, we can probe our semi-supervised low level encoder to enable low level activity recognition. As seen in Table~\ref{tab:results_table}, the EgoCHARM architecture also achieves the highest low level activity recognition performance, with a test F1 score of 0.855 and a test accuracy of 90.64\%, with just 99 additional parameters and 96 additional FLOPs for the probing layer. Our low level activity recognition confusion matrix is shown in Fig.~\ref{fig:ll_cm}. In line with our PCA visualization, Fig~\ref{sec:pca}, there are some misclassifications between more similar classes (e.g., stationary/walking and walking/running), and no errors between the distinct classes (stationary/running).

\subsection{Sensitivity Analysis}
\label{sec:results_sensitivity_analysis}

 \begin{figure*}[htbp]
    \centering
    \includegraphics[width=\textwidth]{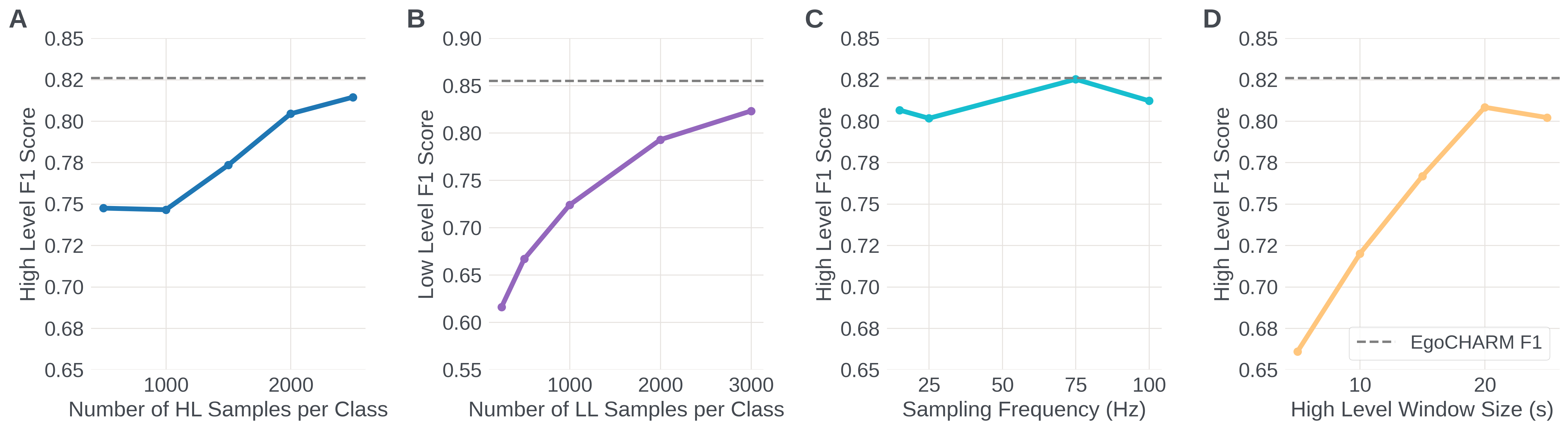}
    \caption{EgoCHARM Sensitivity Analysis. For all plots, the dotted grey line represents the best F1 score for our EgoCHARM model, presented in Table~\ref{tab:results_table}, with all high and low level samples, 50 Hz sampling frequency, and 30s high level window size. \textbf{A}. The effect of the number of 30s samples per class on high level activity classification performance. \textbf{B}. The effect of the number of 1s samples per class on low level activity classification performance. We average results across 4 folds as described in Section~\ref{sec:probing}. \textbf{C}. The effect of IMU sampling frequency on high level activity classification performance. \textbf{D}. The effect of high level window size (s) on high level activity classification performance.}
    \label{fig:ablations}
\end{figure*}

We present results from our sensitivity analysis, detailed in Section~\ref{sec:methods_sensitivity_analysis}, in Fig.~\ref{fig:ablations}. We additionally include the test F1 scores and accuracies in table format in Table~\ref{sup:ablation_num_samples_HL}, Table~\ref{sup:ablation_num_samples_LL}, Table~\ref{sup:ablation_resampling_freq}, and Table~\ref{sup:ablation_window_size}.

\subsubsection{Sample Efficiency}
\label{subsec:sample_efficiency}
Collecting data from human subjects is expensive and time-consuming, making it crucial to understand the number of samples needed for training machine learning algorithms. For high level activity classification, we restrict the number of training samples to 500, 1000, 1500, 2000, 2500 per class and plot the test F1 scores in Fig.~\ref{fig:ablations}A. EgoCHARM achieves reasonable performance of 0.748 F1 score with just 500 samples per class, which equates to only about 83 minutes of data per class, when taking into account the window size of 30s and the stride of 10s for our high level dataset. To achieve 0.814 F1 score, which is close to the best performance of 0.826 F1 score, we need 2500 samples per class, equating to about 416 minutes of data per class. 

Meanwhile, for low level activity classification, we restrict the number of training samples for our probing layer to 250, 500, 1000, 2000, and 3000 per class and plot the test F1 scores using 4 fold cross validation in Fig.~\ref{fig:ablations}B. To achieve 0.823 F1 score, which is comparable to the best performance of 0.855 F1 score, we need only 3000 samples per class, which equates to only 50 minutes of data per class, when taking into account the window size of 1s and stride of 1s for our low level dataset. 

\subsubsection{Sampling Frequency}
\label{subsec:sampling_freq}
We also investigate how sampling frequency affects performance. A lower IMU sampling frequency results in reduced power consumption. Furthermore, decreasing the sampling frequency in raw data architectures lowers the input dimensionality, which subsequently reduces the computational load and frees resources for other tasks on the device. All results presented in Table~\ref{tab:results_table} were obtained using a frequency of 50 Hz. To further explore the effect of sampling frequency on high level activity classification performance, we experimented with sampling frequencies of 15, 25, 75, and 100 Hz using our EgoCHARM architecture and plotted the test F1 scores in Fig.~\ref{fig:ablations}C. The findings indicate that lowering the sampling frequency to 15 and 25 Hz does not affect performance, with test F1 scores of 0.807 and 0.802, respectively, compared to 0.826 for the 50 Hz EgoCHARM model. The FLOPs count for the LL encoder for the 15 Hz model reduces to 295,992 from 1,041,072 for the 50 Hz model. Additionally, the results demonstrate that higher sampling frequencies do not enhance the performance of our EgoCHARM model.

\subsubsection{High Level Window Size}
\label{subsec:hl_window_size}
Finally, we assess how the size of the high level window impacts the performance of high level activity classification. Decreasing the high level window size from 30 seconds could facilitate applications that demand lower latency for high level activity predictions. We tested window sizes of 5, 10, 15, 20, and 25 seconds. Our findings are illustrated in Fig.~\ref{fig:ablations}D. We observe that classification performance increases as the window size is increased. The performance starts to level off around a 20-second window size, achieving a test F1 score of 0.808, compared to a test F1 score of 0.826 with a 30-second window size for our optimal EgoCHARM model.

\section{Discussion}
\begin{figure}[htbp]
    \centering
    \includegraphics[width=0.7\columnwidth]{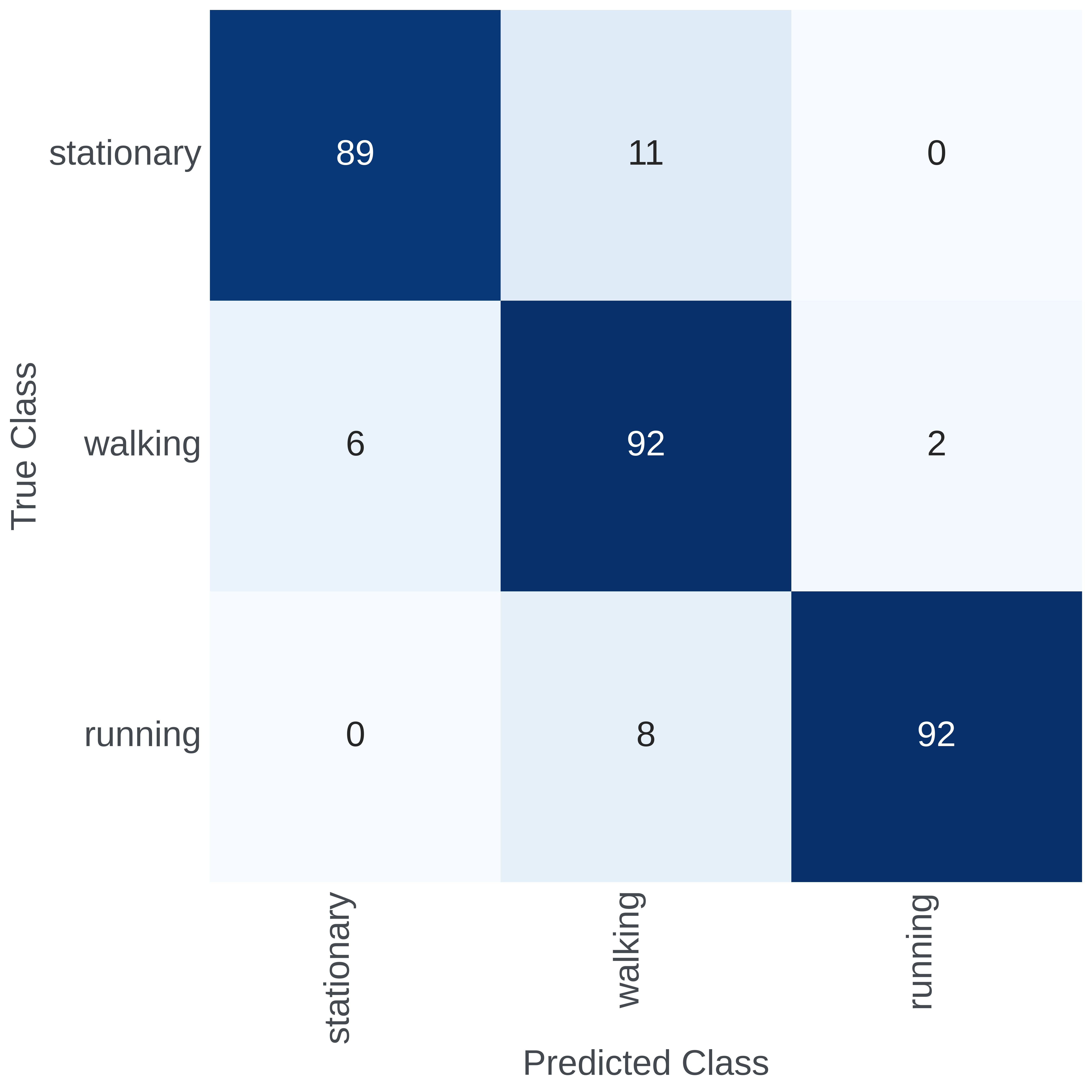}
    \caption{Low Level Activity Recognition Confusion Matrix using EgoCHARM. Values in the confusion matrix are in percentage. We average results across all 4 folds as described in Section~\ref{sec:probing}.}
    \label{fig:ll_cm}
\end{figure}

 In this work, we introduced EgoCHARM, a resource-efficient hierarchical machine learning architecture designed for both high and low level activity recognition using a single egocentric IMU. Our results demonstrated that EgoCHARM achieved high performance metrics for both high and low level activity recognition while maintaining a small number of parameters (22k low level parameters and 63k high level parameters). In comparison, other egocentric IMU architectures, like IMU2CLIP and PRIMUS, use over 1.4 million parameters~\cite{moon2022imu2clip, das2024primus}, for their IMU encoders. Both our low and high level confusion matrices, shown in Fig.~\ref{fig:ll_cm} and Fig.~\ref{fig:hl_cm}, suggest that EgoCHARM performs particularly well in detection of higher motion activities. Our results additionally reveal that our novel low level encoder architecture, shown in Fig.~\ref{fig:ll_encoder}, achieves superior performance in egocentric activity recognition compared to other architectures, including the IMU2CLIP~\cite{moon2022imu2clip} encoder architecture. Notably, we found that varying dilation in the CNN layers plays a crucial role in extracting meaningful feature representations from IMU signals.

Our sensitivity analysis in Section~\ref{sec:results_sensitivity_analysis} offered additional insights into further reducing the resources required by EgoCHARM during both training and deployment. Specifically, in Section~\ref{subsec:sample_efficiency}, we discovered that EgoCHARM maintains competitive performance with a relatively small number of 500 high level samples per class and 3000 low level samples per class, highlighting its cost-effectiveness. In Section~\ref{subsec:sampling_freq}, we showed that decreasing the IMU sampling frequency can further reduce power and computational demands. Specifically, we found that a 15 Hz sampling frequency yields performance comparable to 50 Hz. This reduction results in a substantial 3.5x decrease in FLOPs for our low level encoder, from 1,041,072 to 295,992, making EgoCHARM even more computationally efficient. Finally, in Section~\ref{subsec:hl_window_size}, we found that reducing the high level window size to 20 seconds can decrease latency in activity predictions while maintaining performance similar to 30-second windows.

\begin{figure*}[ht]
    \centering
    \includegraphics[width=0.9\textwidth]{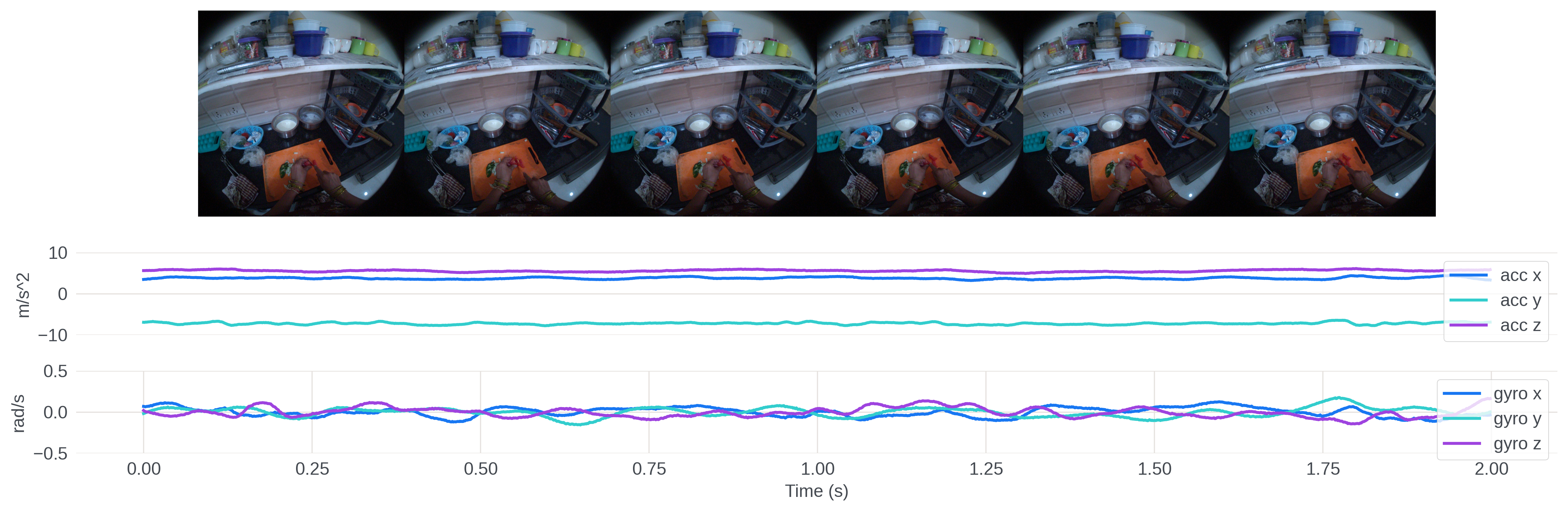}
    \caption{Egocentric camera frames and IMU Signals are shown from a 2 second clip of the cooking high level activity. In this clip, the participant cuts tomatoes on a cutting board using a knife.}
    \label{fig:cooking}
\end{figure*}

\begin{figure*}[ht]
    \centering
    \includegraphics[width=0.9\textwidth]{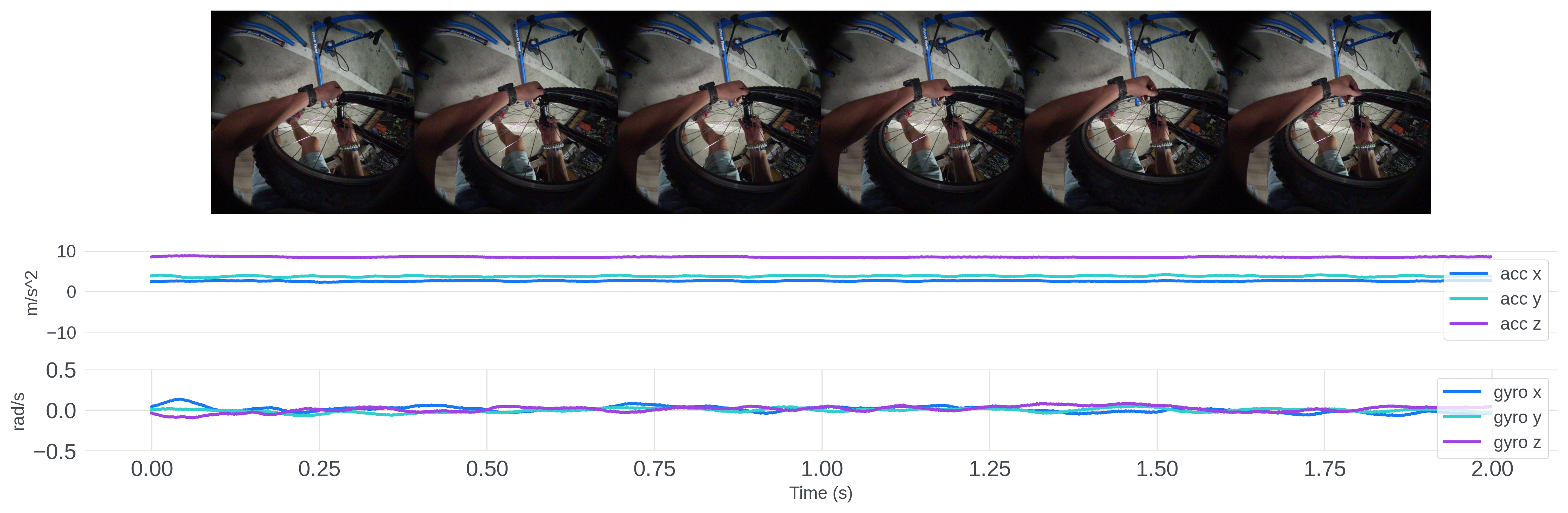}
    \caption{Egocentric camera frames and IMU Signals are shown from a 2 second clip of the bike repair high level activity. In this clip, the participant, while bent over, tightens a nut by hand to fasten the wheel onto the bike frame.}
    \label{fig:bike_repair}
\end{figure*}

\subsection{Advantages of EgoCHARM}
Through our exploration, we find that EgoCHARM offers benefits over other approaches for on-device, always-on activity recognition. Firstly, EgoCHARM achieves high performance compared to many current methods for activity recognition using egocentric IMUs, while requiring less memory and compute. EgoCHARM is also sample efficient, needing only a small number of training samples for both high and low level activities to achieve high performance. 

Secondly, we demonstrated that EgoCHARM's low level encoder can automatically learn low level motion embeddings using only high level labels, as qualitatively shown with PCA in Fig.~\ref{fig:pca_LL}. Utilizing only high level labels in EgoCHARM's training process cuts down on the costs and time involved in data collection and annotation, since annotating low level activities is much more time-intensive than annotating high level ones. We also illustrated how these low level embeddings are generalizable, showing strong performance in low level activity recognition using just a simple probing layer with only 99 parameters. We anticipate that these motion embeddings could be beneficial for other downstream tasks, directly utilized by other models on devices for applications such as image stabilization and providing context to AI assistant models~\cite{moon2024anymal}.

Lastly, the two-stage model architecture of EgoCHARM facilitates the simple division of the algorithm, enabling the low level encoder to be loaded and executed directly on always-on IMU chips equipped with compute capabilities, while the high level architecture exists on the main device processor. This design choice is particularly advantageous as it allows the frequently executed low level architecture to operate entirely independently of the main device processor, minimizing interrupts and freeing up computational and memory resources. 

\subsection{Inherent Limitations of Egocentric IMU Signals}
\label{sec:limitations_egocentric_IMU}
Our high level activity confusion matrix (Fig.~\ref{fig:hl_cm}) shows that EgoCHARM faces challenges in distinguishing activities characterized by low motion and manipulation of objects. While alternative methods and architectures might offer slightly improved performance on these classes, we believe the inherent limitations of raw data from head-mounted, egocentric IMUs hinder the ability to accurately differentiate these activities. As an example, we present raw IMU data from two distinct activities, cooking and bike repair, in Fig.~\ref{fig:cooking} and Fig.~\ref{fig:bike_repair}. In the cooking scenario, the participant cuts tomatoes on a cutting board with a knife, whereas in the bike repair scenario, the participant tightens a nut by hand while bent over, securing the wheel to the bike frame. Despite the significant differences in these activities from both human motion and visual perspectives, the egocentric IMU signals appear almost identical, showing only minor accelerations and angular velocities. The primary distinguishing factor in the motion signals is the participant's bent posture during the bike repair activity, which results in different signal offsets for the three acceleration axes due to the effect of the gravity acceleration vector.

This observation highlights that egocentric IMUs excel at detecting head motion and high-motion activities where the head moves in conjunction with the body. However, they struggle in situations with minimal motion or when the body and arms move independently of the head. To overcome these limitations, it may be necessary to pair egocentric IMUs with other devices or sensing modalities, such as wrist IMUs or egocentric cameras, depending on the downstream application, enhancing sensing capabilities and improving overall performance. Nonetheless, our research demonstrates that egocentric IMUs can be pivotal for always-on, resource-efficient activity recognition and sensor cascading on smartglasses.

\subsection{Opportunities and Future Work}
While the EgoCHARM algorithm shows considerable promise, several avenues for future investigation remain to be explored. Firstly, additional evaluation is necessary with a broader range of classes, both high and low level, to determine how well EgoCHARM can scale while maintaining strong performance. Moreover, to thoroughly assess resource efficiency, especially compute, future efforts should focus on deploying and benchmarking our algorithm directly on-chip. Secondly, our methods were tested using datasets collected from a single device (Aria V1 glasses) and one IMU location (head). However, EgoCHARM could serve as an effective approach for training fully generalizable IMU encoders using only high level labels. To achieve this, further investigation is required to explore generalization across various devices and body locations without increasing memory, computational, or sample demands. Lastly, as discussed in Section~\ref{sec:limitations_egocentric_IMU}, there are inherent limitations in egocentric IMU signals. Future work should explore how egocentric IMU embeddings can be integrated directly with other modalities or with sensor cascading to expand class scalability and enhance performance while remaining resource-efficient.

\section{Conclusion}
In conclusion, we introduced EgoCHARM, a novel hierarchical machine learning algorithm for recognizing both high level and low level activities using a single egocentric IMU on smartglasses. Our resource-efficient algorithm achieves strong performance in high level activity recognition while also demonstrating the utility of our semi-supervised low level encoder in the downstream task of classifying low level activities. By classifying both high and low level activities, we can gain a comprehensive understanding of an individual's daily routine, enabling applications in health monitoring, activity tracking, and personalized recommendations. Our findings highlight the opportunities and limitations of using egocentric, head-mounted IMUs for activity recognition and provide insights into future research directions in this field. With the increasing adoption of consumer smartglasses, our work has implications for the further development of always-on, context-aware AI assistants that can seamlessly integrate into users' daily lives.


\bibliographystyle{abbrv-doi}

\bibliography{bibliography}

\pagebreak
\section*{Supplemental Materials}
\label{sec:supplemental_materials}
\setcounter{section}{0}
\renewcommand{\thesection}{S\arabic{section}}

\setcounter{subsection}{0}
\renewcommand{\thesubsection}{S\arabic{section}.\arabic{subsection}}

\setcounter{subsubsection}{0}
\renewcommand{\thesubsubsection}{S\arabic{section}.\arabic{subsection}.\arabic{subsubsection}}

\setcounter{figure}{0}
\renewcommand{\thefigure}{S\arabic{figure}}

\setcounter{table}{0}
\renewcommand{\thetable}{S\arabic{table}}

\setcounter{equation}{0}
\renewcommand{\theequation}{S\arabic{equation}}

\section{Samples per Activity Class}

\begin{figure}[H]
    \centering
    \includegraphics[width=\columnwidth]{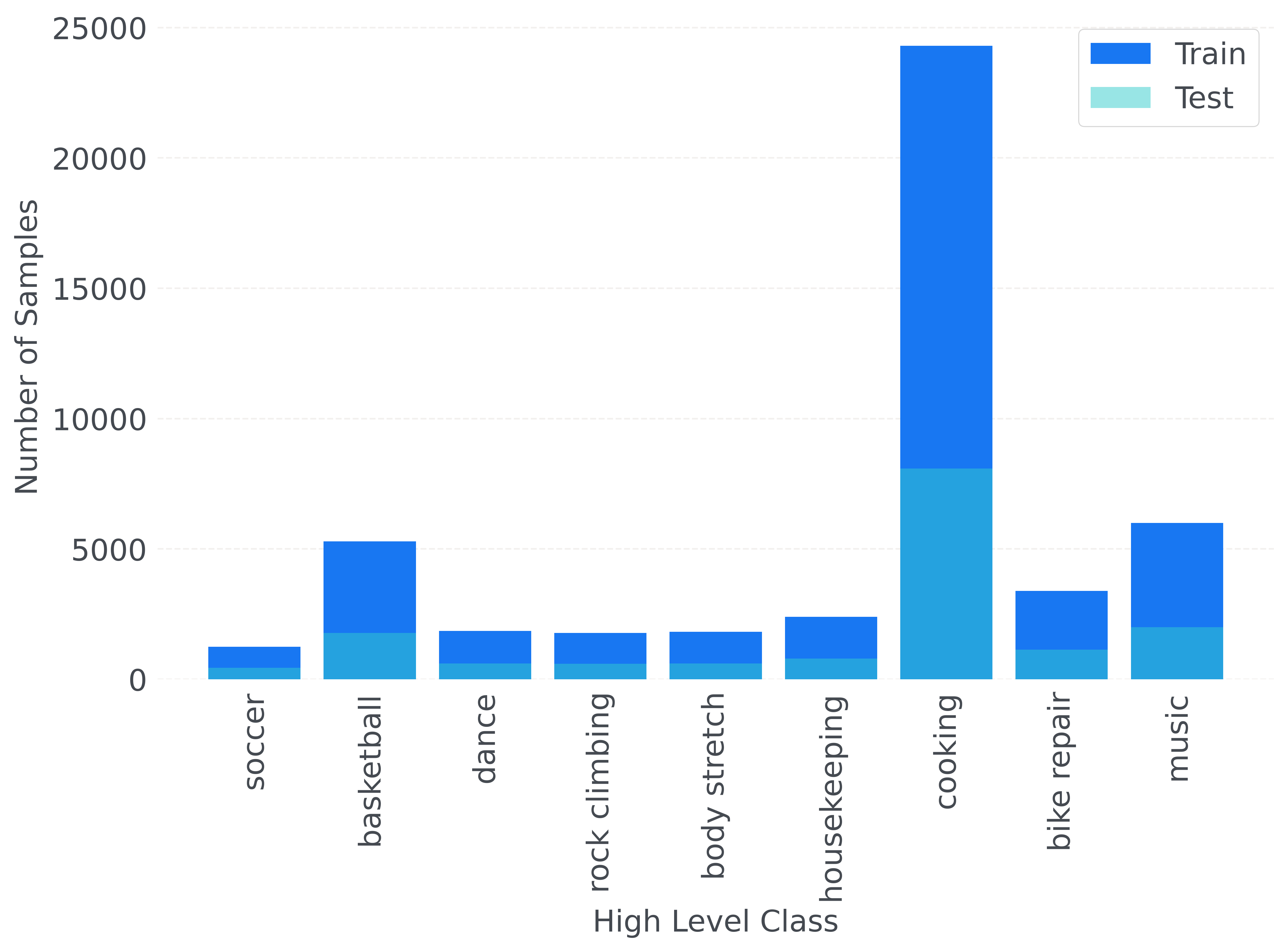}
    \caption{The number of 30 second train and test samples per class for the high level dataset.}
    \label{fig:hl_samples}
\end{figure}

\begin{figure}[H]
    \centering
    \includegraphics[width=\columnwidth]{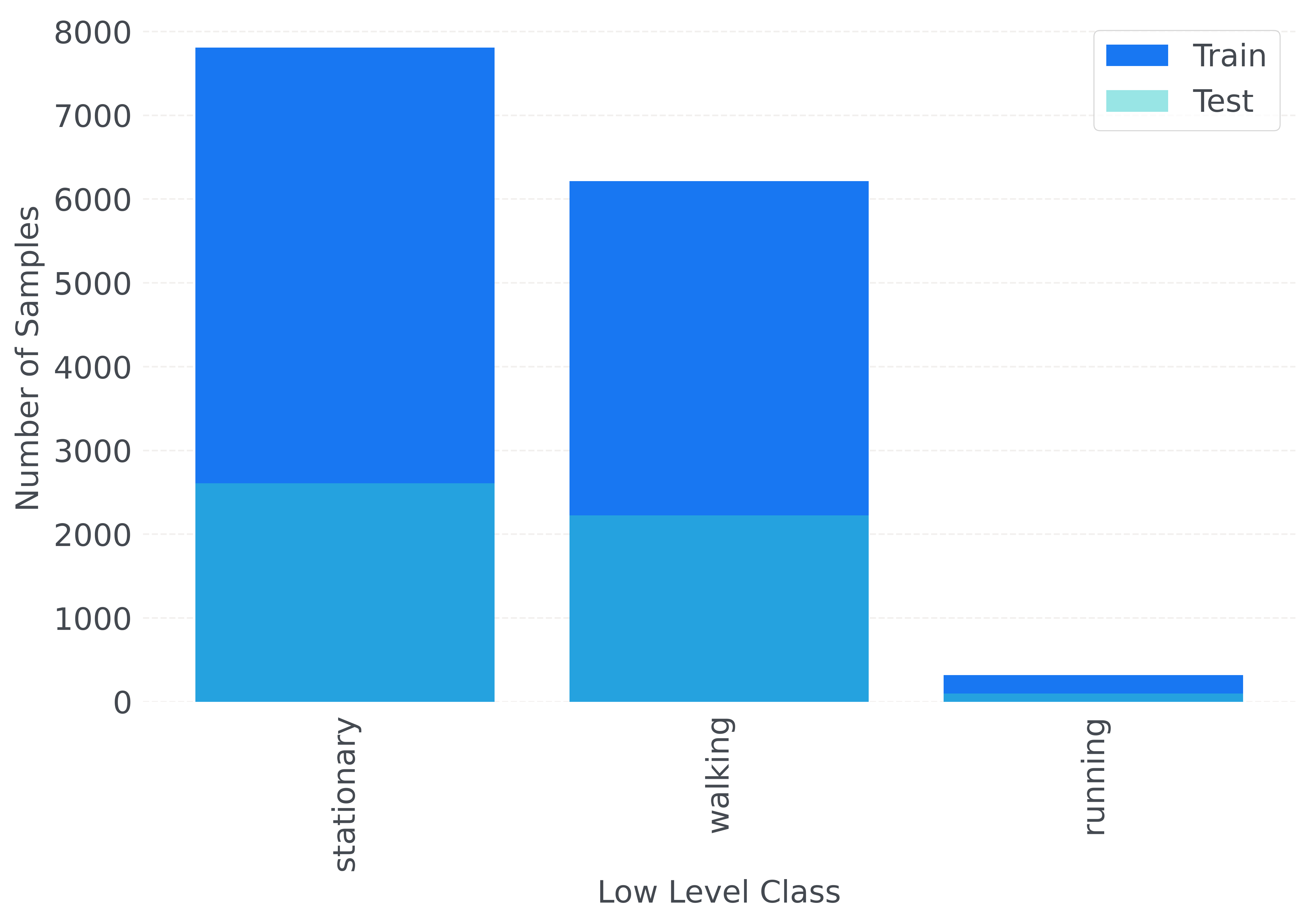}
    \caption{The number of 1 second train and test samples per class for the low level dataset. Due to the small number of running samples, we apply 4 fold stratified cross validation. The data shown in this plot is from a single fold.}
    \label{fig:ll_samples}
\end{figure}

In Fig.~\ref{fig:hl_samples} and Fig.~\ref{fig:ll_samples}, we provide plots with the number of samples for each activity for both our high and low level datasets respectively. 

\section{Hyperparameter Sweeps}
\label{sup:hyperparam_searches}

For our hyperparameter searches involving combinations of high and low level architectures, with results shown in Table~\ref{tab:results_table}, we explore the following low level architecture parameters: maximum dilation, number of output channels per kernel, number of CNN layers, number of GRU/LSTM layers, average versus max pooling, output embedding dimension, and dropout. For the high level architecture parameters, we examine hidden layer dimension, number of hidden layers, and dropout. For our hyperparameter sweeps, we specifically search over values that yield models with a low number of parameters. We employ a step learning rate scheduler for training, spanning 30 total epochs. Regarding learning parameters, we search over learning rate, batch size, gamma (the multiplicative factor for learning rate decay), and step size (the period of learning rate decay).

In our sensitivity analysis hyperparameter sweeps, we keep all other parameters constant while varying learning rate, batch size, maximum epochs, dropout, gamma, and step size.

\section{Sensitivity Analysis Table Results} 
\label{sup:sensitivity_analysis}

Our sensitivity analysis results are presented in table format below. 

\begin{table}[H]
\centering
\caption{Effect of the Number of HL Training Samples per Class on Test F1 and Accuracy for HL Activity Classification}
\vspace{-1em}
\begin{tabular}{ccc}
\toprule
\textbf{Number of Samples} & \textbf{F1 Score} & \textbf{Accuracy (\%)} \\
\midrule
500  & 0.748 & 74.58 \\
1000 & 0.747 & 76.15 \\
1500 & 0.774 & 76.34 \\
2000 & 0.804 & 79.36 \\
2500 & 0.814 & 82.65 \\
\textbf{All} & \textbf{0.826} & \textbf{82.86} \\
\bottomrule
\label{sup:ablation_num_samples_HL}
\end{tabular}
\caption*{Note: The bolded row highlights the specific parameter configuration used in the EgoCHARM algorithm.}
\end{table}

\begin{table}[H]
\centering
\caption{Effect of the Number of LL Training Samples per Class on Test F1 and Accuracy for LL Activity Classification}
\vspace{-1em}
\begin{tabular}{ccc}
\toprule
\textbf{Number of Samples} & \textbf{F1 Score} & \textbf{Accuracy (\%)} \\
\midrule
250  & 0.616 & 71.40 \\
500  & 0.667 & 79.88 \\
1000 & 0.724 & 84.87 \\
2000 & 0.793 & 88.51 \\
3000 & 0.823 & 89.67 \\ 
\textbf{All} & \textbf{0.855} & \textbf{90.64} \\
\bottomrule
\label{sup:ablation_num_samples_LL}
\end{tabular}
\caption*{Note: The bolded row highlights the specific parameter configuration used in the EgoCHARM algorithm.}
\end{table}

\begin{table}[H]
\centering
\caption{Effect of Sampling Frequency on Test F1 and Accuracy for High Level Activity Recognition}
\vspace{-1em}
\begin{tabular}{ccc}
\toprule
\textbf{Sampling Frequency} & \textbf{F1 Score} & \textbf{Accuracy (\%)} \\
\midrule
15  & 0.807 & 82.21 \\
25  & 0.802 & 80.26 \\
\textbf{50} & \textbf{0.826} & \textbf{82.86} \\
75  & 0.825 & 82.86 \\
100 & 0.812 & 81.58 \\
\bottomrule
\label{sup:ablation_resampling_freq}
\end{tabular}
\caption*{Note: The bolded row highlights the specific parameter configuration used in the EgoCHARM algorithm.}
\end{table}

\begin{table}[H]
\centering
\caption{Effect of High Level Window Size on Test F1 and Accuracy for High Level Activity Recognition}
\vspace{-1em}
\begin{tabular}{ccc}
\toprule
\textbf{HL Window Size (s)} & \textbf{F1 Score} & \textbf{Accuracy (\%)} \\
\midrule
5  & 0.661 & 65.62 \\
10 & 0.720 & 70.75 \\
15 & 0.767 & 77.77 \\
20 & 0.808 & 84.11 \\
25 & 0.802 & 79.92 \\
\textbf{30} & \textbf{0.826} & \textbf{82.86} \\
\bottomrule
\label{sup:ablation_window_size}
\end{tabular}
\caption*{Note: The bolded row highlights the specific parameter configuration used in the EgoCHARM algorithm.}
\end{table}

\end{document}